\documentclass{article}
\usepackage{spconf,graphicx}
\usepackage{makecell}
\usepackage{xcolor}
\usepackage{flushend}
\usepackage{cite}
\usepackage{amsmath,amssymb,amsfonts}
\usepackage{algorithmic}
\usepackage{textcomp}
\usepackage{enumitem}
\usepackage{subfigure}
\usepackage{soul}
\usepackage{bm}

\def\EDL{\text{NYCTALE}} 
\def\Q{\bm{Q}}
\def\K{\bm{K}}
\def\V{\bm{V}}
\def\W{\bm{W}}
\title{$\EDL$: Neuro-Evidence Transformer for Adaptive and Personalized Lung Nodule Invasiveness Prediction}

\name{Sadaf Khademi$^{\dag}$,
Anastasia Oikonomou$^{\ddag}$, 
Konstantinos Plataniotis$^{\ddag\ddag}$,
 Arash Mohammadi$^{\dag}$}
\address{$~^{\dag}$Concordia Institute for Information Systems Engineering, Concordia University, Montreal, Canada\\
$~^{\ddag}$Department of Medical Imaging, Sunnybrook Health Sciences Centre,  Toronto, Canada\\
$~^{\ddag\ddag}$Department of Electrical and Computer Engineering, University of Toronto, Toronto, Canada}

\begin{document}
\ninept
\maketitle
\begin{abstract}
Drawing inspiration from the primate brain's intriguing evidence accumulation process, and guided by models from cognitive psychology and neuroscience, the paper introduces the NYCTALE framework, a neuro-inspired and evidence accumulation-based Transformer architecture. The proposed neuro-inspired NYCTALE offers a novel pathway in the domain of Personalized Medicine (PM) for lung cancer diagnosis. In nature, Nyctales are small owls known for their nocturnal behavior, hunting primarily during the darkness of night. The NYCTALE operates in a similarly vigilant manner, i.e., processing data in an evidence-based fashion and making predictions dynamically/adaptively. Distinct from conventional Computed Tomography (CT)-based Deep Learning (DL) models, the NYCTALE performs predictions only when sufficient amount of evidence is accumulated. In other words, instead of processing all or a pre-defined subset of CT slices, for each person, slices are provided one at a time. The NYCTALE  framework then computes an evidence vector associated with contribution of each new CT image. A decision is made once the total accumulated evidence surpasses a specific threshold. Preliminary experimental analyses conducted using a challenging in-house dataset comprising $114$ subjects. The results are noteworthy, suggesting that NYCTALE outperforms the benchmark accuracy even with approximately $60$\% less training data on this demanding and small dataset. 
\end{abstract}

\section{Introduction}
\vspace{-.1in}
The landscape of Personalized Medicine (PM) in oncology stands at a critical crossroads~\cite{piccialli2021survey}, marked by both unprecedented high-risk challenges and transformative high-reward opportunities~\cite{Johnson:2021}. Unlike the traditional approach of focusing solely on tumor features, PM targets tailoring treatments based on tumor, environmental, lifestyle, and/or patient molecular profiles. A paradigm shift has, therefore, swept through the medical image community, moving from traditional disease-focused approaches to data-centric disease identification. Particularly, Deep Neural Networks (DNNs)~\cite{Parnian:SPM, Zhang:2018} are becoming integral in diagnosing disease patterns, tailoring treatment plans, and enhancing prediction accuracy~\cite{Gambardella:2020}. Unlike traditional methods, Deep Learning (DL)-based techniques~\cite{Yang:2022, Papadakis:2019, Schork:2019} can delve into patient-disease correlations, positioning themselves as an indispensable and integral part of future PM paradigm. Capitalizing on these observations, we introduce an innovative and intuitively pleasing DL architecture, the $\EDL$ framework, for lung nodule invasiveness prediction. The $\EDL$ is designed based on concepts from Evidence Accumulation Models (EAMs)~\cite{Ratcliff2016,ratcliff2008diffusion, tanaka2023emerging} within the domain of cognitive psychology and neuroscience. 

The $\EDL$ is particularly designed for lung nodule invasiveness prediction ~\cite{zhang2017radiomics, afshar2019handcrafted, davri2023deep}, specifically, Non-small Cell Lung Cancer (NSCLC), as it remains a leading cause of cancer-related mortality globally~\cite{miller2022cancer}. The World Health Organization (WHO) reports that lung cancer is responsible for approximately $1.8$ million deaths each year~\cite{thandra2021epidemiology}. NSCLC represents the most prevalent form of lung cancer, with Lung Adenocarcinoma (LUAC)~\cite{chen2014non} being the most common subtype. SubSolid Nodules (SSNs), characterized by their combination of Ground-Glass Opacity (GGO) and solid components, differ from purely solid nodules. Studies~\cite{henschke2002ct} have shown that SSNs have a considerably higher malignancy rate compared to solid nodules, highlighting their critical clinical significance in relation to early-stage LUAC. Despite strides in medical advancements, the battle against NSCLC is hampered by late diagnoses and the reality of rapid post-surgery recurrence (poor outcomes). Furthermore, diagnosis of SSNs is crucially challenging as their invasiveness level can significantly influence malignancy likelihood and treatment decisions. Initiating treatment at an early stage can lead to timely PM interventions and development of effective treatment strategies, substantially improving recovery chances and patient survival rates. 

\setlength{\textfloatsep}{0pt}
\begin{figure*}[t!]
\centering
\includegraphics[width=0.9
\linewidth]{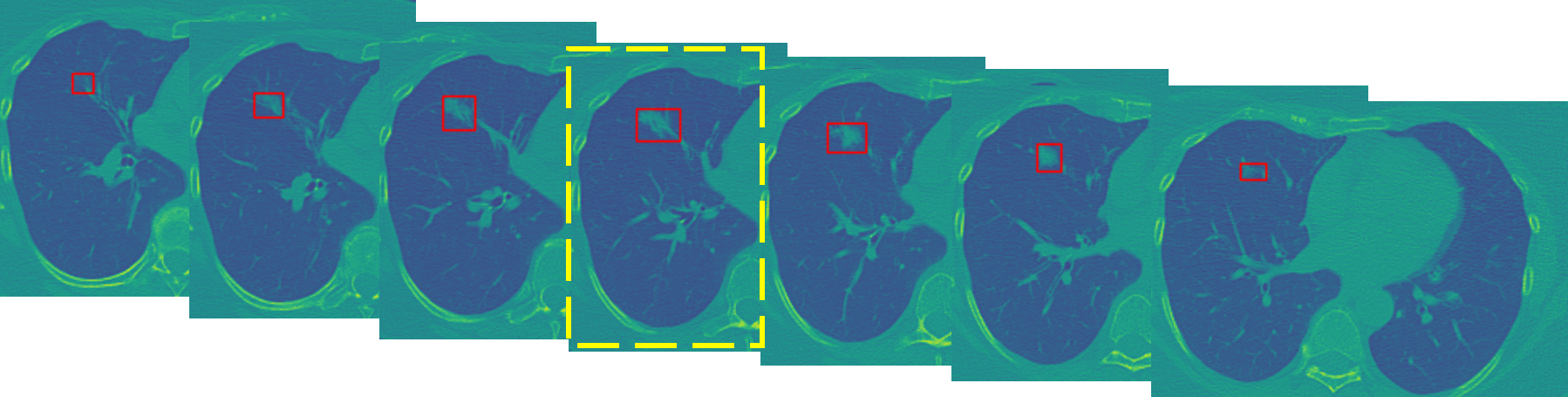}
\vspace{-.1in}
\caption{Sequence of sample slices in a CT volume. The yellow dashed rectangle highlights the middle slice. \label{fig:seq}}
\vspace{-.2in}
\end{figure*}
When it comes to early detection and assessment of SSNs, Computed tomography (CT) is the primary imaging modality~\cite{Afshar:ICASSP2020} offering high spatial resolution and capturing intricate details of lung tissues. When diagnosing lung cancer using CT scans to identify and assess abnormalities (nodules), typically, the middle slice (at times jointly with adjacent slices) is only considered (Fig.~\ref{fig:seq}), capturing a central view of the tumor. On the positive side, analyzing the middle slice and its adjacent one can quickly provide a snapshot of the lung's condition, enabling faster preliminary assessments. However, there are limitations to this method. Relying solely on the middle slice and its immediate neighbors may overlook significant pathological findings present in slices further away, potentially leading to incomplete diagnostic information. This technique assumes uniformity in the distribution of lesions, which is not always the case. Moreover, this method might not fully leverage the volumetric data provided by CT scans, which can offer a three-dimensional perspective critical for accurate diagnosis and treatment planning. More specifically, lung cancer lesions, especially in the early stages, may be small and located at the periphery of the lungs. Focusing solely on the middle slice and adjacent slices could miss these peripheral lesions, leading to a delayed diagnosis. This heterogeneity requires a comprehensive analysis of the entire lung volume for accurate detection and characterization. Finally, the spread of lung cancer, including metastases to lymph nodes or other lung sections, may not be visible in the central slices, necessitating processing of the entire CT series for accurate diagnosis.

\vspace{.05in}
\noindent
\textbf{Contributions:} In our previous works, we focused on development of different DL-based architectures including but not limited to: (i) Capsule Network  architectures~\cite{sabour2017dynamic} for the task of lung nodule malignancy prediction. Along this path, we have developed different models including, but not limited to, mixture of capsule networks (MIXCAP)~\cite{Parnian:mixcaps}, and the Bayesian capsule networks (BayesCap)~\cite{Parnian:BayesCap}, to 3D multi-scale capsule networks~\cite{Parnian:SREP}, and; (ii)  Transformer architectures~\cite{vaswani2017attention} for the task of lung nodule invasiveness prediction~\cite{Sadaf:DSP23, Sadaf:ICASSP23}. Along this path,  cross-attention is utilized in~\cite{Sadaf:DSP23} to fuse hierarchical features generated by Shifting Window (SWin) Transformer~\cite{liu2021swin} with various hand-crafted radiomics and quantitative measurements of nodule characteristics~\cite{Sadaf:DSP23}. In~\cite{Sadaf:ICASSP23}, we have further advanced our hybrid malignancy predictive framework by integrating spatial and temporal features extracted by two parallel self-attention mechanisms, i.e.,  Convolutional Auto-Encoder (CAE) and SWin Transformer. 

In this study, we focus on an alternative but intuitively pleasing design inspired by the decision-making process of primate brain and recent works~\cite{Katia:2018EAM, Katia:ICHMS2024} on its adaptation to the domain of DL networks. With its extensive and intricate network of neurons, human brain represents the most advanced signal processing system known, capable of analyzing and integrating multiple signals from various modalities simultaneously in an adaptive and real-time fashion. When it comes to decision-making, brain relies on the gradual integration of uncertain or incomplete information over time, accumulating evidence until a certain threshold is reached, thereby determining the precision and timing of responses. Essentially, the neural circuitry weighs different potential actions by processing inconsistent information from various cortical sources.  Drawing inspiration from the aforementioned aspect of primate decision-making and recent works, we aim to create and assess a neuro-inspired transformer model for efficient processing and learning from CT sequences for lung nodule invasiveness prediction. Distinct from conventional DL models that produce an output for every input or after a predetermined sequence of inputs, the proposed $\EDL$ is designed to emit an output once it meets a defined evidence threshold from a continuous flow of CT slices. In other words, instead of feeding the model with all or a pre-defined subset of CT slices, for each subject, individual slices are provided one at a time. The $\EDL$ framework computes a vector indicating the evidence level each observation contributes towards a potential decision. This evidence is then accumulated, and a decision is made once the total evidence surpasses a specific threshold.  We hypothesize that the proposed $\EDL$ not only enhances the computational efficiency of CT-based lung nodule invasiveness prediction but also has the potential to improve interpretability of the PM pipeline. In summary, the paper makes the following key contributions:
\begin{itemize}[noitemsep,topsep=0pt]
\item \textit{\textbf{Neuro-Inspired Transformer Framework:} }
Unlike conventional DL models that require a fixed sequence of inputs or generate an output for each input, $\EDL$ is uniquely designed to produce a decision once a predefined evidence threshold is met, simulating the behavior found in cognitive psychology and neuroscience. This approach allows for a more adaptive and personalized analysis of CT sequences.
\item \textit{\textbf{Enhanced Computational Efficiency for Lung Nodule Diagnosis:}} By leveraging a continuous flow of CT slices and accumulating evidence until a certain threshold is reached, $\EDL$ promises to significantly reduce the computational overhead of lung nodule invasiveness prediction. This methodology addresses the limitations of current practices by ensuring comprehensive examination beyond the middle slice and its immediate neighbors in an adaptive, dynamic, and on need bases only fashion.
\end{itemize}
The rest of paper is organized as follows: Section~\ref{Sec:RWs} briefly outlines the relevant literature. The proposed $\EDL$ framework is presented in Section~\ref{Sec:EDL}. Section~\ref{Sec:Exp} outlines our experimental results based on an in-house dataset. Finally Section~\ref{Sec:Conc} concludes the paper.
\section{Related Works} \label{Sec:RWs}
\vspace{-.1in}

In this section, we briefly overview the related works under two categories of DL-based lung nodule invasiveness prediction, and evidence accumulation modeling followed by a concluding remark.

\vspace{.1in}
\noindent
\textbf{\textit{DL for Lung Nodule Invasiveness Prediction:}} Generally speaking, the existing research on evaluating the invasiveness of SSNs falls into three main categories: (i) Radiomics approaches~\cite{Chen2018, Xu2019} that focus on extracting and analyzing quantitative descriptors of a nodule's shape, texture, and internal complexity; (ii) DL-based architectures ~\cite{Shen2021} that automatically identify and learn relevant patterns directly from imaging data. Recently, DL-based methodologies ~\cite{Li2022,Wang2022} have shown promising results in improving the accuracy and efficiency of SSN diagnosis. DL-models can autonomously identify subtle imaging features and patterns that may be difficult for expert radiologists to detect, and;
(iii) Hybrid methods~\cite{Mehta2021} that combine both strategies addressing the challenges inherent in using either strategy in isolation, such as the requirement for extensive datasets with uniform, pathology-confirmed annotations. 
Volumetric CT scans are made up of a sequence of 2D images, which together form the 3D volume of the lung tissue. In Reference~\cite{Li2019}, for instance, a fusion approach is developed integrating handcrafted features with those learned from a 3D Convolutional Neural Network (CNN) to predict lung nodule malignancy. While 3D-based models can potentially improve performance by extracting reacher spatio-temporal features, such models require large training datasets and are computationally demanding. Consequently, there has been a surge of recent interest~\cite{Farhangi2020} in using sequential DL models for simultaneous analyzing of 2D CT images. For example, Reference~\cite{CARL:2023} focused on accurate classification of histological subtype of NSCLC using CT aiming to assist treatment planning. Targeting a multi-view approach, authors introduced Cross-Aligned Representation Learning (CARL) to jointly earn view-specific and view-invariant representations. Conversely, when DL models are applied to analyze CT scans, it is imperative to focus on key components/regions that are essential for accurate diagnosis. This is where Transformer architectures using attention mechanism~\cite{vaswani2017attention}, in particular, Vision Transformer (ViT)~\cite{Dosovitskiy2020} can play a key role allowing selective focus on specific regions of CT images. Transformer models are capable of capturing long-range dependencies in the input sequence more effectively than their Recurrent Neural Networks (RNNs) and CNNs counterparts~\cite{Paul2022}. In $\EDL$, we adopt a transformer variant, the Shifting Window (SWin) Transformer~\cite{Liu_2021_ICCV}, as the evidence encoder.

\vspace{.05in}
\noindent
\textbf{Evidence Accumulation Models (EAMs):}
Within the domain of cognitive psychology and neuroscience, Evidence Accumulation Models (EAMs)~\cite{evans2019evidence} stand as a pivotal framework elucidating the processes underlying decision-making in both human and animal brains. Central to these models is the concept that decisions are made by the incremental gathering and assessment of information until a critical threshold is reached, prompting a definitive choice. This approach is encapsulated in the Drift Diffusion Model (DDM)~\cite{ratcliff2008diffusion, bogacz2006physics}, a prominent EAM, which simplifies decision-making to a one-dimensional process of noise-driven evidence accumulation towards one of two boundaries, each representing a distinct decision. The DDM, by quantifying the rate of evidence accumulation (drift rate) and the amount of evidence required to make a decision (decision boundary), has profoundly influenced our understanding of the temporal dynamics of decision-making. It offers insightful explanations for the variability in response time and accuracy, illustrating how decisions are influenced by the speed-accuracy trade-off. Moreover, its applicability across a range of cognitive tasks, from simple perceptual decisions to more complex judgments, underscores its versatility and foundational role in cognitive science.

Further advancements in EAMs have led to the development of more explanatory models, such as the Linear Ballistic Accumulator (LBA)~\cite{BROWN2008153}, which accommodate the complexity of decision-making in real-world scenarios. For instance, recently~\cite{Venugopal:2023} LBA model is integrated with Reinforcement Learning (RL) for modeling human behaviour in the Grid-Sailing Task. These models extend the basic premise of DDM by incorporating multiple accumulators, each representing different choices, and allowing for a more efficient representation of how decisions are made when faced with more than two alternatives. Notably, EAMs have been integrated with neuroimaging techniques to identify the neural correlates of evidence accumulation, revealing a distributed network of brain regions~\cite{pisauro2017neural}, including the prefrontal cortex and basal ganglia, that dynamically interact during decision-making. This integration of theoretical models with empirical data has not only validated the concept of evidence accumulation but also enriched our understanding of the neural mechanisms that govern decision-making. Furthermore, the application of EAMs in understanding psychiatric and neurological disorders~\cite{tanaka2023emerging} characterized by impaired decision-making, such as schizophrenia and Parkinson’s disease, highlights their potential in clinical neuroscience, offering pathways to novel therapeutic strategies. Thus, evidence accumulation models continue to provide a robust framework for dissecting the complexities of cognitive functions, bridging the gap between theoretical constructs and biological reality.

Despite their significant contributions to understanding decision-making processes, EAMs have yet to be fully integrated into the domain of advanced DL models. This gap highlights a critical area for interdisciplinary research, where the principles of cognitive psychology and neuroscience could enhance the interpretability and decision-making capabilities of DNNs. The complexity of DNNs, characterized by their high-dimensional, non-linear processing capabilities, contrasts with the relatively straightforward, linear dynamics of traditional EAMs such as the DDM. This discrepancy underscores the need for novel computational approaches that can bridge the simplicity of EAMs with the complexity of DL systems. As highlighted by Wang \textit{et al.}~\cite{wang2018using}, incorporating cognitive models into DL could not only improve the performance of Artificial Intelligence (AI) systems in tasks requiring refined decision-making but also provide new avenues for understanding the brain's decision-making processes through the lens of AI. 
Agrawal, Sycara, and colleagues notably pioneered research in this domain by exploring the Cortico-Basal Ganglia-Thalamic (CBGT) circuits within the brain to inform the design of DL networks, as detailed in their initial work~\cite{Katia:2018EAM}. This exploration led to the creation of an innovative CBGT-inspired network, which was initially based on a shallow encoder design. Building upon this foundational work, Sharma, Sycara \textit{et al.} recently introduced an advancement with the development of the CBGT-Net~\cite{Katia:ICHMS2024}, which incorporates a CNN encoder, demonstrating promising outcomes on benchmark datasets such as MNIST and CIFAR-10.

\vspace{.05in}
\noindent
\textbf{Concluding Remarks:} Capitalizing on the above discussion, we aim to integrate evidence accumulation concept into DL domain for lung nodule Invasiveness prediction. Achieving this objective involves processing extracted features in such a way to reflect evidence accumulation for decision-making. In this regard, the model's predictions across different classes (logits) can be used and interpreted as the strength of the model's evidence for each class. 
In our context of integrating EAMs with DL, the logits serve as a measure of evidence for or against each decision option. Since logits are proportional to the model's confidence in each class, they can be directly used to represent the ``evidence'' that is being accumulated. The adaptively weighted evidence accumulator is initialized at zero or some baseline value representing initial uncertainty. CT slices are then processed one by one, where for each slice the logits are extracted and used to update the evidence accumulators. This process simulates the accumulation of evidence over time or iterations. For the task of lung nodule invasiveness prediction using CT sequences, which naturally unfold over time, logits from each CT image can be used to update the accumulators, reflecting how evidence evolves. 
Integrating a DDM-like evidence accumulation process using logits offers a complex and delicate approach to decision-making in DL models, allowing them to make more informed and dynamic decisions based on the accumulation of evidence over time. Our findings demonstrate enhancement through the utilization of an evidence-based accumulation technique. Significantly, this improvement is achieved while requiring much less training data (about $60\%$ in our experiments). This outcome underscores the efficiency of the proposed $\EDL$ framework, offering promising implications for the optimization of decision-making processes in healthcare domain.
\vspace{-.15in}
\section{The $\EDL$ Framework} \label{Sec:EDL}
\vspace{-.1in}

\begin{figure*}[t] 
\centerline{\includegraphics[width=0.85\linewidth]{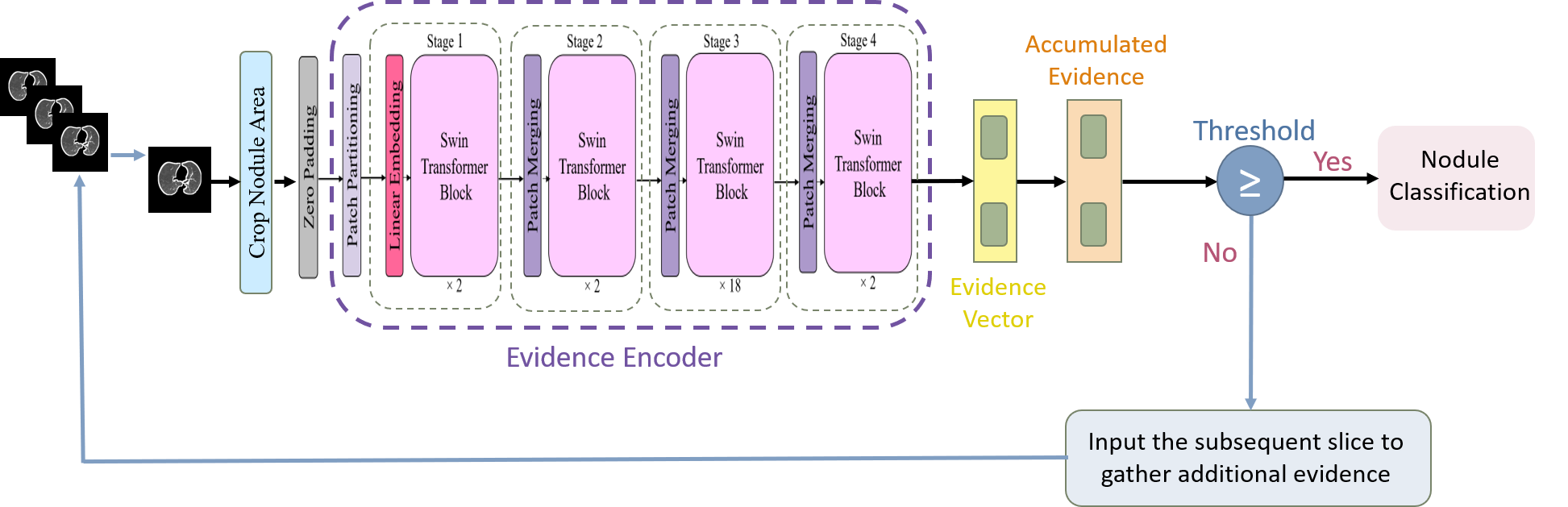}}
\vspace{-.2in}
\caption{\footnotesize Architecture of the $\EDL$ framework. \label{fig}}
\vspace{-.15in}
\end{figure*}

In this section, we present details of the proposed $\EDL$ framework, which composed of two main components: an Evidence Encoder and an Accumulation Module. As stated previously, the evidence encoder utilizes a SWin transformer-based architecture to extract evidence vectors from CT slices. Each evidence vector is then processed through the accumulation module to evaluate the available evidence for decision-making. In what follows, initially, we will outline the data structure and its preparation process. Following that, we will delve into the evidence encoder, which serves as the feature extraction procedure, detailing how it generates the desired input for the evidence module. Finally, we will discuss the evidence accumulation technique and the classification mechanism.

\subsection{Data Description}
In this study, we utilized the in-house dataset introduced in Reference~\cite{Ana:2019}, supplemented with five additional cases from the same institution to improve data balance. This dataset comprises non-thin volumetric CT scans of $114$ SSNs confirmed as part of the adenocarcinoma spectrum which were further classified into three groups based on pathological findings including pre-invasive, minimally invasive, and invasive adenocarcinomas. CT scans were collected using technical specifications ranging from $100$ to $135$ kVp and $80$ to $120$ mAs. For the purpose of our research, we selected the most recent scan taken before the resection date. Consistent with the approach taken in the original study, we combined the pre-invasive and minimally invasive categories to convert the multi-class problem into a binary one, resulting in $58$ non-invasive and $56$ invasive cases. 

\subsection{Data Preparation}
Within each patient's CT volume, the slices containing visible nodules were evaluated and chosen independently by two radiologists who were not informed of the pathology findings. The chosen slices were subsequently passed to a pre-processing phase to prepare the input structure for the evidence encoder. The initial segmented region surrounding each SSN was produced using the commercial software Vitrea v$7.3$. This software employs image processing algorithms~\cite{Puskar:2017} to precisely segment nodules in CT images. Following Vitrea's process, radiologists reviewed the automatically segmented regions and manually adjusted the contours as necessary to ensure more precise annotations.

\subsection{Evidence Encoder}
The proposed evidence encoder is designed according to a hierarchical attention-based structure to discover informative spatial features from nodule patches. The overall architecture of our evidence encoder including its four processing stages is shown in Fig.~\ref{fig}. At first, the segmented area surrounding each nodule resulting from the annotation coordinates in pre-processing step is zero-padded to ($224,224$) pixels and divided into $4\times4$ patches. In Stage $1$, the linear embedding layer projects each patch's feature dimension into $128$ and then SWin Transformer blocks are applied to the features of each patch. The hierarchical feature maps are achieved by the patch merging technique, which is the process of combining the representations generated for multiple patches of the segmented nodule.  This technique concatenates extracted features of neighboring patches and enables the model to capture both local and global information from the input image. The SWin Transformer block is designed according to the original transformer encoder architecture, which uses a self-attention mechanism to capture dependencies among different instances of each segmented nodule in a sequence~\cite{vaswani2017attention}. In self-attention mechanism, input is transformed into three different representations, i.e., the Query ($\Q$), Key ($\K$), and Value ($\V$) with dimensions $d_k$, $d_k$, and $d_v$, respectively. These representations are used to compute attention scores as follows
\begin{equation}
Attention(\Q,\K,\V) = \text{softmax}\left(\frac{\Q\K^{T}}{\sqrt{d_k}}\right)\V,  \label{eq:attention}
\end{equation}
which are then applied to weigh the elements of the input data and generate a weighted representation. In Eq.~\eqref{eq:attention},  superscript $^{T}$ indicates the transpose of a given matrix. For each CT image, attention scores can be computed $h$ times in parallel with various linear projections implemented each by a different ``head'' to provide more informative presentations of data in a process known as Multi-head Self-Attention (MSA). The multiple attention scores are then concatenated and processed by a feed-forward neural network to produce the final attention weights. The output of the MSA is  given~by
\begin{eqnarray}
MSA(\Q,\K,\V) &\!\!\!\!=\!\!\!\!& Concat\big(head_1, \cdots, head_h\big) \W^O, \nonumber \\
\text{where } \quad head_i &\!\!\!\!=\!\!\!\!& Attention(\Q\W_i^Q,\K\W_i^K,\V\W_i^V).\label{eq:mha}
\end{eqnarray}
where the projections are achieved by parameter matrices $\W_i^Q \in \mathbb{R}^{d_{model}\times d_k}$, $\W_i^K \in \mathbb{R}^{d_{model}\times d_k}$, $\W_i^V \in \mathbb{R}^{d_{model}\times d_v}$, and $\W^O \in \mathbb{R}^{{hd_v}\times d_{model}}$.
In each SWin Transformer block there are two consecutive transformer encoder units, named Window MSA (W-MSA) and Shifted Window MSA (SW-MSA) modules replacing the original one (MSA) to calculate self-attention within local windows. The W-MSA module computes attention within non-overlapping windows of patches, where each window has a size of $7\times 7$ patches. In the following unit, the SW-MSA module calculates self-attention in shifted W-MSA windows. Each window is shifted by half its size, allowing the SW-MSA module to capture overlapping features between adjacent windows.  In this study, we implemented the SWin-B architecture of SWin transformer that has $2$, $2$, $18$, and $2$ layers for Stages $1$ to $4$, respectively.  Due to the small size of our dataset, we fine-tuned a pre-trained SWin-B transformer trained on ImageNet-21k dataset~\cite{Imagenet:2009}, which provides a feature map consisting of $1,024$ features for each patient. The output feature vectors generated for the slices of each subject from Stage $4$ were passed to the Evidence Accumulation module, which is described next.

\subsection{Evidence Accumulation Module}
In the evidence accumulation module, we transform the feature vector obtained from our evidence encoder through a Fully Connected (FC) layer into a two-element evidence vector representing logit values for each class. The accumulated evidence consists of a vector that tracks the total evidence gathered from the beginning of the input stream, initialized with a value of zero.
To delve deeper, the accumulated evidence is updated through a DDM-like accumulation process to calculate a weighted average of acquired evidences. The weights are proportional to the uncertainty of the evidence vectors. As stated previously, DDM, as a cognitive model that describes how decisions are made under conditions of uncertainty, is based on the concept of evidence accumulation, where information supporting different choices is gathered over time until enough evidence is accumulated to cross a decision threshold. An important hyperparameter, known as the evidence threshold, plays a crucial role in determining whether the necessary evidence has been acquired. This value is optimized through a nested cross-validation process. The threshold acts as the determinant for decision-making. If any accumulated evidence surpasses this threshold, the accumulated values are directed to a softmax function for classification. Otherwise, the next CT slice is processed by the evidence encoder to gather further evidence for the network.

\vspace{-.15in}
\section{Experimental Results} \label{Sec:Exp}
\vspace{-.1in}
\begin{table*}[t!]
\caption{\footnotesize Classification performance (mean$\pm$std) of different aggregation methods in DL models. \label{tab:agg} }
\vspace{.1in}
\centering
\begin{tabular}{|c|c|c|c|c|c|}
\hline
Model &  Accuracy (\%) & Sensitivity (\%) & Specificity (\%) & AUC & Number of slices\\
\hline
CNN (Voting)~\cite{DL:2023}& 60.36$\pm$8.14 & 67.31$\pm$16.43 & 51.47$\pm$24.19 & 0.49$\pm$0.17 & 834$\pm$13\\
\hline
ViT (Voting)~\cite{Vit:2023} & 61.97$\pm$13.75& 72.00$\pm$14.53 & 49.36$\pm$17.22 & 0.54$\pm$0.15 &834$\pm$13\\
\hline
SWin (GAP) & 61.66$\pm$16.46& 45.66$\pm$32.07 & 77.99$\pm$24.73 & 0.68$\pm$0.16&834$\pm$13\\
\hline
SWin (GMP) &78.10$\pm$12.45& 76.66$\pm$21.80 & 79.66$\pm$17.91 & \textbf{0.80$\pm$0.15} &834$\pm$13\\
\hline
$\EDL$ &\textbf{81.66$\pm$11.09}& \textbf{79.00$\pm$14.37} & \textbf{83.34$\pm$22.36} & 0.77$\pm$0.12 & \textbf{245$\pm$108}\\
\hline
\end{tabular}
\vspace{-.2in}
\end{table*}
In this section, we evaluate the performance of the $\EDL$ framework by employing a  $10$ fold nested cross-validation technique. Nested cross-validation is a robust technique used to evaluate model performance and select optimal hyperparameters. It involves two levels of cross-validation, i.e., the outer loop and the inner loop. In the former, the dataset is divided into training and validation sets multiple times to assess the model's performance, while in the latter, each training set is further divided into subsets for hyperparameter tuning. The entire $\EDL$ network underwent training in an end-to-end fashion. We fine-tuned the pre-trained SWin-B transformer on our in-house dataset using the AdamW optimizer. This training phase utilized a learning rate of $1e-5$, a weight decay of $0.05$, and was conducted over $50$ epochs, with an early stopping training strategy employed, set to halt training if no improvement was observed in validation loss for $10$ consecutive epochs. During the training process, each subject's slices were sequentially passed through the SWin transformer to encode features. This iterative procedure persisted until reaching the predefined evidence threshold for any element in the accumulated evidence vector. Subsequently, the loss was calculated based on the predicted label for the subject's CT volume using a cross-entropy loss function. Below, various scenarios are investigated to evaluate nodule invasiveness prediction performance of the $\EDL$ framework from different aspects.

\vspace{.1in}
\noindent
\textbf{4.1. Comparison of Subject-based Aggregation Methods}
\vspace{.05in}

\noindent
As stated previously, there is a significant challenge associated with identifying the most valuable and informative slice within CT volumes. Additionally, one of the difficulties encountered in processing CT volumes among different subjects is the inconsistency in the number of slices where nodules are visible. Consequently, certain studies employ various aggregation methods to leverage all available slices and combine predictions from individual slices into a single prediction for each subject, a practice that may lead to increased computational costs. Some common aggregation methods include: {\textit{(i) Voting}}: In this method, predictions of each slice are considered as votes. The final prediction for the subject is determined by the most frequent prediction among all slices. Voting is straightforward and easy to implement, however, it may not take into account the confidence or uncertainty associated with each prediction. {\textit{(ii) Global Max Pooling (GMP)}}: In global max pooling, the maximum feature value across all slices is selected as the final feature for the subject. This method emphasizes the highest feature value among all slices. {\textit{(iii) Global Average Pooling (GAP)}}: In average pooling, the features of all slices are averaged to compute the final feature vector for the subject. This method considers the collective information from all slices and provides a more balanced approach compared to voting or global max pooling. It can help mitigate the impact of outlier values associated with individual slices. 
We applied the aforementioned aggregation methodologies on well-known DL models such as CNN and ViT (using voting) and Swin Transformer (with GMP and GAP layers) to compare their performance on our in-house dataset. The classification results are presented in Table~\ref{tab:agg}. As it can be observed from Table~\ref{tab:agg}, GMP acts as the best aggregation selection approach. 

\vspace{.05in}
\noindent
\textbf{4.2. Effect of Threshold Value}
\begin{figure}[t!]
\centering
\includegraphics[width=0.8
\linewidth]{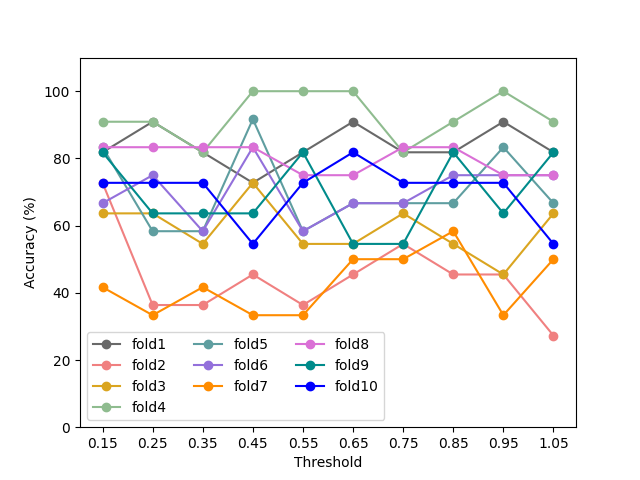}
\vspace{-.2in}
\caption{\footnotesize Model performance across different threshold values with slice addition direction of left to right. \label{fig:thR}}
\vspace{-.28in}
\end{figure}
\begin{figure}[t!]
\centering
\mbox{\subfigure[]{\includegraphics[scale=0.28]{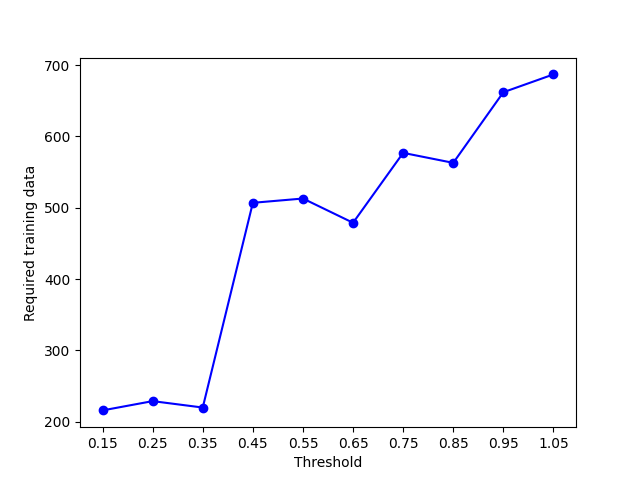}}
\subfigure[]{\includegraphics[scale=0.28]{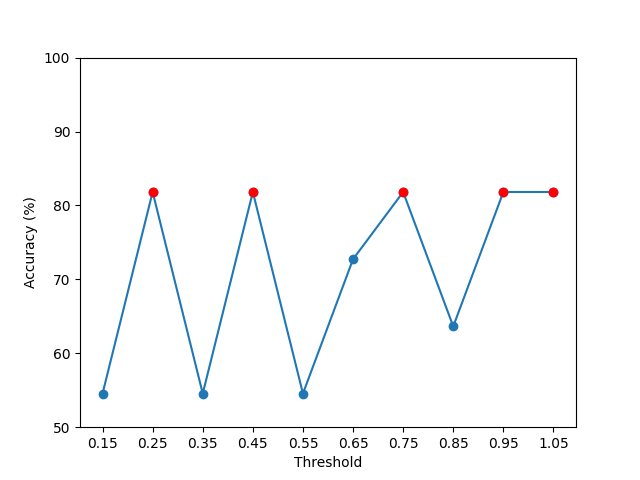}}}
\vspace{-.2in}
\caption{\footnotesize (a) Amount of training data (total number of selected slices in fold $1$) needed in the last epoch for different threshold values. (b) Model performance (fold $1$) across different threshold values with slice addition direction of middle to the sides. \label{fig:train}\label{fig:thM}}
\end{figure}

\noindent
\textbf{\textit{Framework Performance:}} The threshold variable as a hyperparameter in our model, is optimized during model training-validation phase within a specified range, from $0.05$ to $1.05$, with an increment step of $0.05$. However, we aim to visualize the impact of threshold variation on our classification task. Fig.~\ref{fig:thR} depicts how the classification accuracy varies with different threshold values for $10$ folds. For instance, model attained its peak accuracy for fold $1$ at multiple threshold values ($0.25$, $0.65$, and $0.95$), matching the accuracy achieved by Swin-GMP, identified as the best-performing aggregation-based DL.

\begin{figure}[t!]
\begin{minipage}[b]{.48\linewidth}
  \centering
  \centerline{\includegraphics[width=4.8cm]{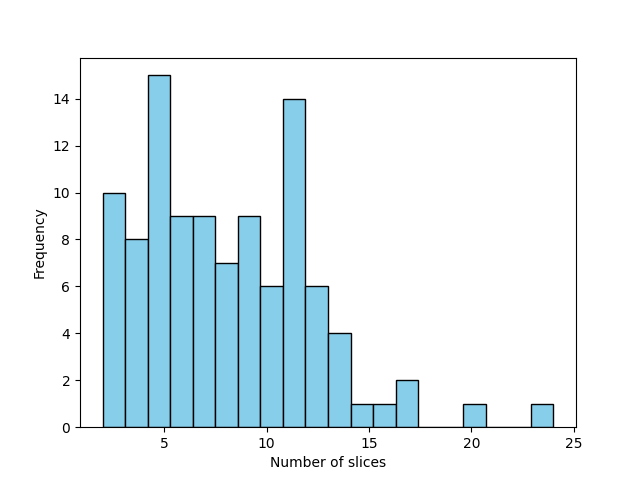}}
  \centerline{(a)}\medskip
\end{minipage}
\hfill
\begin{minipage}[b]{0.48\linewidth}
  \centering
  \centerline{\includegraphics[width=4.8cm]{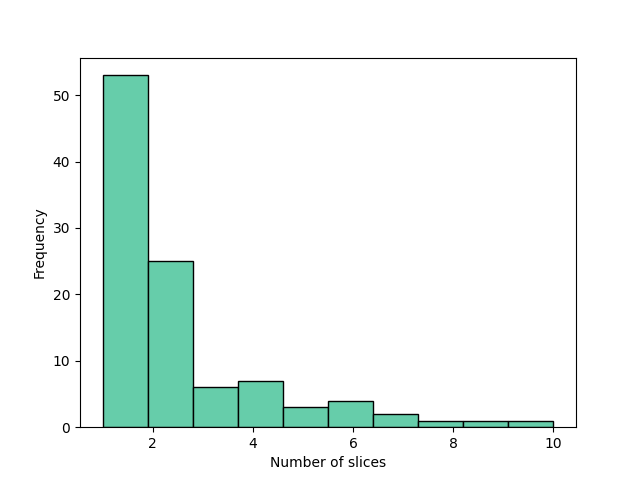}}
  \centerline{(b)}\medskip
\end{minipage}
\vspace{-.25in}
\caption{\footnotesize Histogram of training slices in fold $1$ across various subjects. (a) Distribution of slice numbers in the original training dataset, (b) Distribution of selected slices used for the last epoch of the training phase \label{fig:res}.}
%
\end{figure}

\noindent
\textbf{\textit{Required Number of Slices for Training:}} Fig.~\ref{fig:train}(a) illustrates the required amount of fold $1$ training data for the $\EDL$ framework based on varying threshold values. As expected, elevating the threshold value leads to an increased number of slices to surpass the threshold. Consequently, the optimal threshold for fold $1$ remains at $0.25$, ensuring the minimal number of slices necessary to attain the desired performance level. It is worth noting that in comparison to Swin-GMP, the $\EDL$ framework demonstrates comparable accuracy levels but achieves this in an adaptive manner fitted to each subject, utilizing significantly less training data. Specifically, the Swin-GMP model used $834$ slices on average in each epoch of the training process while in our framework, each epoch utilizes a much smaller number of slices. Fig.~\ref{fig:res} shows the variation in slice numbers within the training set of data acquired by the $\EDL$ framework for the last epoch of training. As can be observed from Fig.~\ref{fig:res}, only $213$ slices are employed by our model.

\vspace{.1in}
\noindent
\textbf{4.3. Effect of Slice Addition Directionality}

\noindent
As previously mentioned, selecting the optimal slice within a CT volume for processing and classification poses significant challenges. However, as evident from Fig.~\ref{fig:seq}, it is apparent that the middle slice provides a more comprehensive view of the nodule, which is chosen in most single-slice processing studies. In this study, we employed two approaches for accumulating slices. The first approach (left to right) begins from the first slice of the volume and adds additional slices if there is not sufficient evidence for decision-making. Conversely, the second approach (middle to the sides) starts from the middle slice and progressively adds neighboring slices. Fig.~\ref{fig:thR} illustrates the results of the first approach, while Fig.~\ref{fig:thM}(b) depicts the outcomes of the second approach for fold $1$. It can be inferred that evidence accumulation from the growth of the nodule is more effective, as observed in the first approach.

\vspace{-.15in}
\section{Conclusion} \label{Sec:Conc}
\vspace{-.1in}
In this paper, we introduced an innovative and intuitively pleasing DL architecture, the $\EDL$ framework, for lung nodule invasiveness prediction. The proposed $\EDL$ framework represents an advancement in the application of DL within the evolving domain of PM in oncology. By mimicking the human brain's approach to decision-making through the gradual integration of uncertain information, $\EDL$ sets a new standard for diagnosing NSCLC. This model's ability to process and learn from CT sequences in a manner that closely resembles natural cognitive processes could revolutionize how medical professionals approach the early detection and treatment planning of lung cancer, aligning with the goals of PM. Through its adaptive and subject-specific analysis, $\EDL$ promises improved computational efficiency and accuracy, embodying a significant leap towards integrating cognitive insights into DL models for healthcare.

\footnotesize
\bibliographystyle{IEEEtran}

\end{document}